\documentstyle[psfig,aclap]{article}
\title{Aggregate and mixed-order Markov models for \\
statistical language processing}
\author{Lawrence Saul and Fernando Pereira \\
{\tt \{lsaul,pereira\}@research.att.com} \\
AT\&T Labs -- Research \\
180 Park Ave, D--130 \\
Florham Park, NJ 07932}
\date{}

\newlength{\listwidth}
\newcommand{\wordlist}[1]{\parbox{\listwidth}{\begin{list}{}{%
\setlength{\leftmargin}{0pt}%
\setlength{\rightmargin}{0pt}\setlength{\topsep}{2pt}%
\setlength{\partopsep}{0pt}}%
\raggedright\item[]#1\end{list}}}
\newcommand{\myquote}[1]{$\langle\mbox{#1}\rangle$}

\begin{document}

\maketitle

\begin{abstract}
We consider the use of language models whose size and accuracy are
intermediate between different order $n$-gram models.  Two types of
models are studied in particular.  Aggregate Markov models are
class-based bigram models in which the mapping from words to classes
is probabilistic.  Mixed-order Markov models combine bigram models
whose predictions are conditioned on different words.  Both types of
models are trained by Expectation-Maximization (EM) algorithms for
maximum likelihood estimation.  We examine smoothing procedures in
which these models are interposed between different order $n$-grams.
This is found to significantly reduce the perplexity of unseen word
combinations.
\end{abstract}

\section{Introduction}
The purpose of a statistical language model is to assign high
probabilities to likely word sequences and low probabilities to
unlikely ones.  The challenge here arises from the combinatorially
large number of possibilities, only a fraction of which can ever be
observed.  In general, language models must learn to recognize word
sequences that are functionally similar but lexically distinct.  The
learning problem, one of generalizing from sparse data, is
particularly acute for large-sized vocabularies (Jelinek, Mercer, and
Roukos, 1992).

The simplest models of natural language are $n$-gram Markov models.
In these models, the probability of each word depends on the $n-1$
words that precede it.  The problems in estimating robust models of
this form are well-documented.  The number of parameters---or
transition probabilities---scales as~$V^{n}$, where $V$ is the
vocabulary size.  For typical models (e.g., $n=3$, $V=10^4$), this
number exceeds by many orders of magnitude the total number of words
in any feasible training corpus.

The transition probabilities in $n$-gram models are estimated from the
counts of word combinations in the training corpus.  Maximum
likelihood (ML) estimation leads to zero-valued probabilities for
unseen $n$-grams.  In practice, one adjusts or {\it smoothes} (Chen and
Goodman, 1996) the ML estimates so that the language model can
generalize to new phrases.  Smoothing can be done in many ways---for
example, by introducing artificial counts, backing off to lower-order
models (Katz, 1987), or combining models by interpolation
(Jelinek and Mercer, 1980).

Often a great deal of information is lost in the smoothing procedure.
This is due to the great discrepancy between $n$-gram models of
different order.  The goal of this paper is to investigate models that
are intermediate, in both size and accuracy, between different order
$n$-gram models.  We show that such models can ``intervene'' between
different order $n$-grams in the smoothing procedure.  Experimentally,
we find that this significantly reduces the perplexity of unseen word
combinations.

The language models in this paper were evaluated on the ARPA North 
American Business News (NAB) corpus.  All our experiments used a 
vocabulary of sixty-thousand words, including tokens for punctuation, 
sentence boundaries, and an unknown word token standing for all 
out-of-vocabulary words.  The training data consisted of approximately 
78 million words (three million sentences); the test data, 13 million 
words (one-half million sentences).  All sentences were drawn randomly 
without replacement from the NAB corpus.  All perplexity figures given 
in the paper are computed by combining {\em sentence} probabilities; 
the probability of sentence $w_{0}w_{1}\cdots w_{n}w_{n+1}$ is given 
by $\prod_{i=1}^{n+1} P(w_{i}|w_{0}\cdots w_{i-1})$, where $w_{0}$ and 
$w_{n+1}$ are the start- and end-of-sentence markers, respectively.  
Though not reported below, we also confirmed that the results did not 
vary significantly for different randomly drawn test sets of the same 
size.

The organization of this paper is as follows.  In 
Section~\ref{sAggregate}, we examine {\it aggregate} Markov models, or 
class-based bigram models (Brown {\em et al.}, 1992) in which the 
mapping from words to classes is probabilistic.  We describe an 
iterative algorithm for discovering ``soft'' word classes, based on 
the Expectation-Maximization (EM) procedure for maximum likelihood 
estimation (Dempster, Laird, and Rubin, 1977).  Several features make 
this algorithm attractive for large-vocabulary language modeling: it 
has no tuning parameters, converges monotonically in the 
log-likelihood, and handles probabilistic constraints in a natural 
way.  The number of classes,~$C$, can be small or large depending on 
the constraints of the modeler.  Varying the number of classes leads 
to models that are intermediate between unigram ($C=1$) and bigram 
($C=V$) models.

In Section~\ref{sMix}, we examine another sort of ``intermediate'' 
model, one that arises from combinations of non-adjacent words.  
Language models using such combinations have been proposed by Huang 
{\em et al.} (1993), Ney, Essen, and Kneser (1994), and Rosenfeld (1996), 
among others.  We consider specifically the {\em skip-$k$} transition 
matrices, $M(w_{t-k},w_{t})$, whose predictions are conditioned on the 
$k$th previous word in the sentence.  (The value of $k$ determines how 
many words one ``skips'' back to make the prediction.)  These 
predictions, conditioned on only a single previous word in the 
sentence, are inherently weaker than those conditioned on all $k$ 
previous words.  Nevertheless, by combining several predictions of 
this form (for different values of $k$), we can create a model that is 
intermediate in size and accuracy between bigram and trigram models.

{\it\mbox{Mixed-order}} Markov models express the predictions
$P(w_t|w_{t-1},w_{t-2},\ldots, w_{t-m})$ as a convex combination of
skip-$k$ transition matrices, $M(w_{t-k},w_t)$.  We derive an EM
algorithm to learn the mixing coefficients, as well as the elements of
the transition matrices.  The number of transition probabilities in
these models scales as $mV^2$, as opposed to $V^{m+1}$.  Mixed-order
models are not as powerful as trigram models, but they can make much
stronger predictions than bigram models.  The reason is that quite
often the immediately preceding word has less predictive value than
earlier words in the same sentence.

In Section~\ref{sSmooth}, we use aggregate and mixed-order models to 
improve the probability estimates from $n$-grams.  This is done by 
interposing these models between different order $n$-grams in the 
smoothing procedure.  We compare our results to a baseline trigram 
model that backs off to bigram and unigram models.  The use of 
``intermediate'' models is found to reduce the perplexity of unseen 
word combinations by over 50\%.

In Section~\ref{sDiscussion}, we discuss some extensions to these 
models and some open problems for future research.  We conclude that 
aggregate and mixed-order models provide a compelling alternative to 
language models based exclusively on $n$-grams.

\section{Aggregate Markov models}
\label{sAggregate}
In this section we consider how to construct class-based bigram models
(Brown {\em et al.}, 1992).  The problem is naturally formulated as one of
hidden variable density estimation.  Let $P(c|w_1)$ denote the
probability that word $w_1$ is mapped into class~$c$.  Likewise, let
$P(w_2|c)$ denote the probability that words in class $c$ are followed
by the word $w_2$.  The class-based bigram model predicts that word
$w_1$ is followed by word $w_2$ with probability
\begin{equation}
\label{eAggregate}
P(w_2|w_1) = \sum_{c=1}^{C} P(w_2|c) P(c|w_1),
\end{equation}
where $C$ is the total number of classes.  The hidden variable in this
problem is the class label $c$, which is unknown for each word $w_1$.
Note that eq.~(\ref{eAggregate}) represents the $V^2$ elements of the
transition matrix $P(w_2|w_1)$ in terms of the $2CV$ elements of
$P(w_2|c)$ and $P(c|w_1)$.

The Expectation-Maximization (EM)~algorithm (Demp\-ster, Laird, and
Rubin, 1977) is an iterative procedure for estimating the parameters
of hidden variable models.  Each iteration consists of two steps: an
E-step which computes statistics over the hidden variables, and an
M-step which updates the parameters to reflect these statistics.

The EM algorithm for aggregate Markov models is particularly simple.
The E-step is to compute, for each bigram $w_1 w_2$ in the training
set, the {\it posterior} probability
\begin{equation}
\label{eAggregatePosterior}
P(c|w_1,w_2) = \frac{P(w_2|c) P(c|w_1)}{\sum_{c'} P(w_2|c') P(c'|w_1)}.
\end{equation}
Eq.~(\ref{eAggregatePosterior}) gives the probability that word $w_1$
was assigned to class $c$, based on the observation that it was
followed by word $w_2$.  The M-step uses these posterior probabilities
to re-estimate the model parameters.  The updates for aggregate Markov
models are:
\begin{eqnarray}
P(c|w_1) & \leftarrow &
\frac{\sum_{w} N(w_1,w) P(c|w_1,w)}
     {\sum_{wc'} N(w_1,w) P(c'|w_1,w)}, \\
P(w_2|c) & \leftarrow &
\frac{\sum_{w} N(w,w_2) P(c|w,w_2)}
     {\sum_{ww'} N(w,w') P(c|w,w')},
\end{eqnarray}
where $N(w_1,w_2)$ denotes the number of counts of $w_1w_2$ in the
training set.  These updates are guaranteed to increase the overall
log-likelihood,
\begin{equation}
\label{eAggregateLogLikelihood}
\ell = \sum_{w_1 w_2} N(w_1,w_2) \ln P(w_2|w_1),
\end{equation}
at each iteration.  In general, they converge to a local (though not
global) maximum of the log-likelihood.  The perplexity $V^{*}$ is
related to the log-likelihood by \mbox{$V^{*} = e^{-\ell/N}$}, where
$N$ is the total number of words processed.

\begin{figure*}
\begin{center}
\begin{tabular}{c@{\hspace{0.5in}}c}
\mbox{\psfig{figure=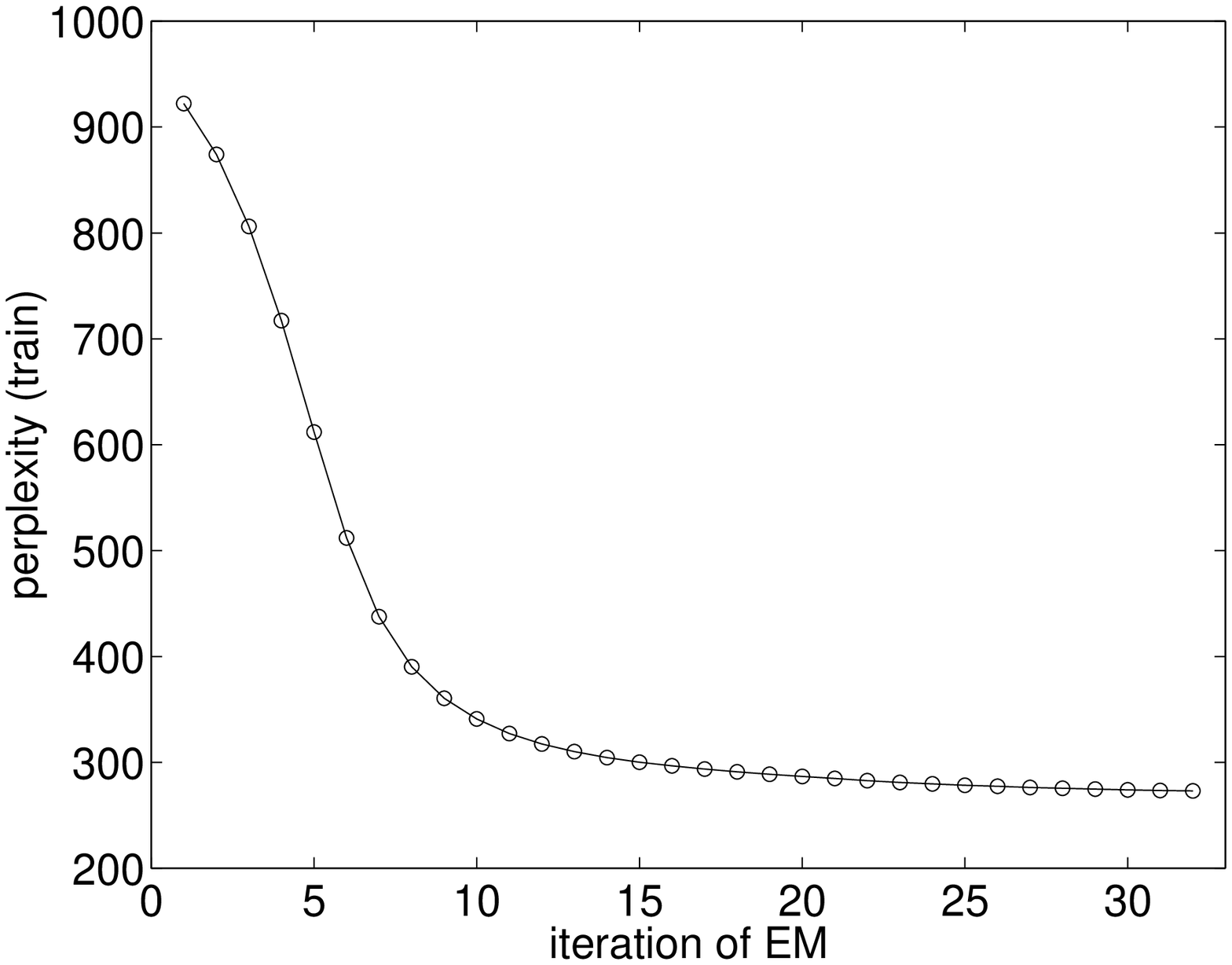,width=2in}} &
\mbox{\psfig{figure=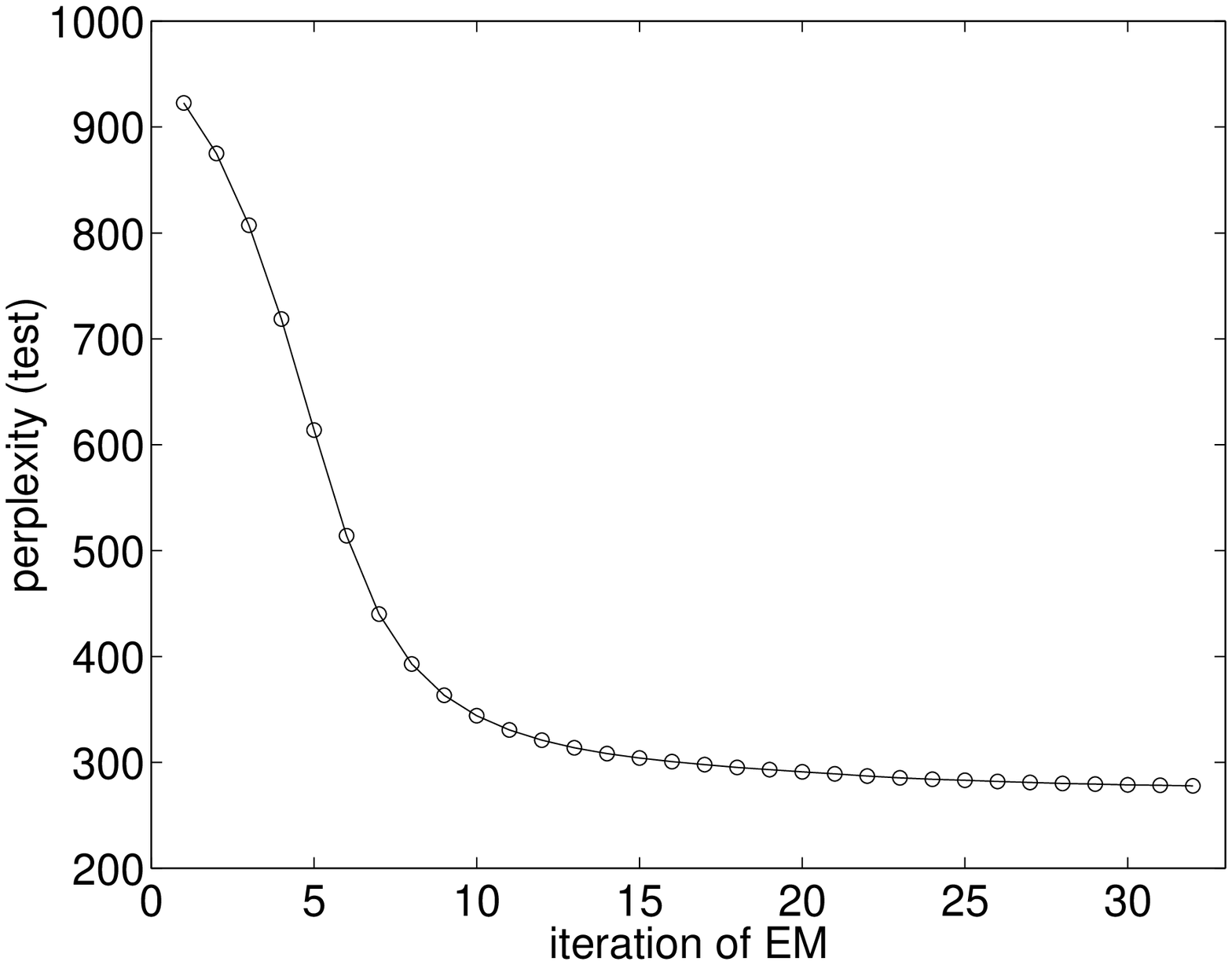,width=2in}} \\
(a) & (b)
\end{tabular}
\end{center}
\caption{Plots of (a) training and (b) test perplexity versus number
of iterations of the EM algorithm, for the aggregate Markov model with
$C=32$ classes.}
\label{fAggregateEM}
\end{figure*}

\begin{table}
\begin{center}
\begin{tabular}{|c|c|c|}
\hline C & train & test \\ \hline 1 & 964.7 & 964.9 \\ 2 & 771.2 &
772.2 \\ 4 & 541.9 & 543.6 \\ 8 & 399.5 & 401.5 \\ 16 & 328.8 & 331.8
\\ 32 & 278.9 & 283.2 \\ V & 123.6 & --- \\ \hline
\end{tabular}
\end{center}
\caption{Perplexities of aggregate Markov models on the training and
test sets; $C$ is the number of classes.  The case $C=1$ corresponds
to a ML unigram model; \mbox{$C=V$}, to a ML bigram model.}
\label{tAggregatePerplexity}
\end{table}

\begin{figure}
\centerline{\hbox{\psfig{figure=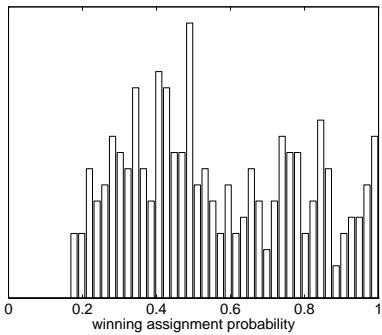,width=2in}}}
\caption{Histogram of the winning assignment probabilities,
$\max_{c} P(c|w)$, for the three hundred most commonly
occurring words.}
\label{fHistogram}
\end{figure}

Though several algorithms (Brown {\em et al.}, 1992; Pereira, Tishby, 
and Lee,
1993) have been proposed for performing the decomposition in
eq.~(\ref{eAggregate}), it is worth noting that only the EM algorithm
directly optimizes the log-likelihood in
eq.~(\ref{eAggregateLogLikelihood}).  This has obvious advantages if
the goal of finding word classes is to improve the perplexity of a
language model.  The EM algorithm also handles probabilistic
constraints in a natural way, allowing words to belong to more than
one class if this increases the overall likelihood.

Our approach differs in important ways from the use of hidden Markov
models (HMMs) for class-based language modeling (Jelinek {\em et al.},
1992).  While HMMs also use hidden variables to represent word
classes, the dynamics are fundamentally different.  In HMMs, the
hidden state at time $t+1$ is predicted (via the state transition
matrix) from the hidden state at time $t$.  On the other hand, in
aggregate Markov models, the hidden state at time $t+1$ is predicted
(via the matrix $P(c_{t+1}|w_t)$) from the {\it word} at time $t$.
The state-to-state versus word-to-state dynamics lead to different
learning algorithms.  For example, the Baum-Welch algorithm for HMMs
requires forward and backward passes through each training sentence,
while the EM algorithm we use does not.

We trained aggregate Markov models with 2, 4, 8, 16, and 32 classes.
Figure~\ref{fAggregateEM} shows typical plots of the training and test
set perplexities versus the number of iterations of the EM
algorithm. Clearly, the two curves are very close, and the monotonic
decrease in test set perplexity strongly suggests little if any
overfitting, at least when the number of classes is small compared to
the number of words in the vocabulary.
Table~\ref{tAggregatePerplexity} shows the final perplexities (after
thirty-two iterations of EM) for various aggregate Markov models.
These results confirm that aggregate Markov models are intermediate in
accuracy between unigram ($C=1$) and bigram ($C=V$) models.

\begin{table*}
\begin{center}
{\footnotesize
\setlength{\listwidth}{200pt}
\begin{tabular}{c@{\hspace{5pt}}c}
\begin{tabular}[t]{|@{\hspace{4pt}}r@{\hspace{4pt}}|@{\hspace{2pt}}l|}
 \hline
1 & \wordlist{as cents made make take} \\ \hline
2 & \wordlist{ ago day earlier Friday Monday month quarter reported said
Thursday trading Tuesday Wednesday \myquote{\ldots} } \\ \hline
3 & \wordlist{ even get to} \\ \hline
4 & \wordlist{based days down home months up work years \myquote{\%}} \\ \hline
5 & \wordlist{ those \myquote{,} \myquote{---} } \\ \hline
6 & \wordlist{ \myquote{.} \myquote{?} } \\ \hline
7 & \wordlist{ eighty fifty forty ninety seventy sixty
thirty twenty  \myquote{(} \myquote{$\cdot$} } \\ \hline
8 & \wordlist{ can could may should to will would } \\ \hline
9 & \wordlist{ about at just only or than \myquote{\&} \myquote{;} } \\ \hline
10 & \wordlist{ economic high interest much no such tax united well } \\ \hline
11 & \wordlist{ president } \\ \hline
12 & \wordlist{ because do how if most say so then think very what when where } \\ \hline
13 & \wordlist{according back expected going him plan used way} \\ \hline
15 & \wordlist{ don't I people they we you} \\ \hline
16 & \wordlist{ Bush company court department more officials police retort
spokesman } \\ \hline
17 & \wordlist{ former the } \\ \hline
18 & \wordlist{ American big city federal general house  military
national party political state union York } \\ \hline
\end{tabular} &
\begin{tabular}[t]{|@{\hspace{4pt}}r@{\hspace{4pt}}|@{\hspace{2pt}}l|}
\hline 
19 & \wordlist{ billion hundred million nineteen } \\ \hline
20 & \wordlist{ did \myquote{"} \myquote{'} } \\ \hline
21 & \wordlist{ but called San \myquote{:} \myquote{start-of-sentence} } \\ \hline
22 & \wordlist{ bank board chairman end group members number office out part
percent price prices rate sales shares use } \\ \hline
23 & \wordlist{ a an another any dollar each first good her his its my old
our their this } \\ \hline
24 & \wordlist{ long Mr. year } \\ \hline
25 & \wordlist{ business California case companies
corporation dollars incorporated industry law money
thousand time today war week \myquote{)} \myquote{unknown}} \\ \hline
26 & \wordlist{ also government he it market she that there which who } \\ \hline
27 & \wordlist{ A.  B.  C.  D.  E.  F.  G.  I.  L.
M.  N.  P.  R. S.  T.  U. } \\ \hline
28 & \wordlist{ both foreign international major many new oil other some Soviet
stock these west world } \\ \hline
29 & \wordlist{ after all among and before between by during for from in
including into like of off on over since through told under
until while with } \\ \hline
30 & \wordlist{ eight fifteen five four half last next nine oh one
second seven several six ten third three twelve two zero
\myquote{-} } \\ \hline
31 & \wordlist{ are be been being had has have is it's not still was were } \\ \hline
32 & \wordlist{ chief exchange news public service trade } \\ \hline
\end{tabular}
\end{tabular}
}
\end{center}
\caption{Most probable assignments for the 300 most frequent words in
an aggregate Markov model with $C=32$ classes. Class 14 is absent
because it is not the most probable class for any of the selected
words.)}
\label{tCluster}
\end{table*}

The aggregate Markov models were also observed to discover meaningful
word classes.  Table~\ref{tCluster} shows, for the aggregate model
with $C=32$ classes, the most probable class assignments of the three
hundred most commonly occurring words.  To be precise, for each class
$c^{*}$, we have listed the words for which $c^{*} = \arg\max_{c}
P(c|w)$.  Figure~\ref{fHistogram} shows a histogram of the winning
assignment probabilities, $\max_{c} P(c|w)$, for these words.  Note
that the winning assignment probabilities are distributed broadly over
the interval $[\frac{1}{C},1]$.  This demonstrates the utility of
allowing ``soft'' membership classes: for most words, the maximum
likelihood estimates of $P(c|w)$ do not correspond to a
winner-take-all assignment, and therefore any method that assigns each
word to a single class (``hard'' clustering), such as those used by
Brown {\em et al.} (1992) or Ney, Essen, and Kneser (1994), would lose
information.

We conclude this section with some final comments on overfitting.  Our 
models were trained by thirty-two iterations of EM, allowing for 
nearly complete convergence in the log-likelihood.  Moreover, we did 
not implement any flooring constraints\footnote{It is worth noting, in 
this regard, that individual zeros in the matrices $P(w_2|c)$ and 
$P(c|w_1)$ do not necessarily give rise to zeros in the matrix 
$P(w_2|w_1)$, as computed from eq.~(\ref{eAggregate}).} on the 
probabilities $P(c|w_1)$ or $P(w_2|c)$.  Nevertheless, in all our 
experiments, the ML aggregate Markov models assigned non-zero 
probability to all the bigrams in the test set.  This suggests that 
for large vocabularies there is a useful regime $1\ll C\ll V$ 
in which aggregate models do not suffer much from overfitting.  In this 
regime, aggregate models can be relied upon to compute the 
probabilities of unseen word combinations.  We will return to this 
point in Section~4, when we consider how to smooth $n$-gram language 
models.

\section{Mixed-order Markov models}
\label{sMix}
One of the drawbacks of $n$-gram models is that their size grows
rapidly with their order.  In this section, we consider how to make
predictions based on a convex combination of pairwise correlations.
This leads to language models whose size grows {\it linearly} in the
number of words used for each prediction.

For each $k>0$, the {\em skip}-$k$~transition~matrix $M(w_{t-k},w_{t})$
predicts the current word from the $k$th previous word in the
sentence.  A {\it mixed-order} Markov model combines the information
in these matrices for different values of~$k$.  Let $m$ denote the
number of bigram models being combined.  The probability distribution
for these models has the form:
\begin{eqnarray}
\label{eMixedOrderModel}
\lefteqn{P(w_{t}|w_{t-1},\ldots,w_{t-m})=} \\
& & \sum_{k=1}^m \lambda_k(w_{t-k})\;M_k(w_{t-k},w_{t})
\prod_{j=1}^{k-1}[1-\lambda_j(w_{t-j})].\nonumber
\end{eqnarray}
The terms in this equation have a simple interpretation.  The $V\times 
V$ matrices $M_k(w,w')$ in eq.~(\ref{eMixedOrderModel}) define the 
skip-$k$ stochastic dependency of $w'$ at some position $t$ on $w$ at 
position $t-k$; the parameters $\lambda_k(w)$ are mixing coefficients 
that weight the predictions from these different dependencies.  The 
value of $\lambda_k(w)$ can be interpreted as the probability that the 
model, upon seeing the word $w_{t-k}$, looks no further back to make 
its prediction (Singer, 1996).  Thus the model predicts from $w_{t-1}$ 
with probability $\lambda_1(w_{t-1})$, from $w_{t-2}$ with probability 
$[1-\lambda_1(w_{t-1})]\lambda_2(w_{t-2})$, and so on.  Though 
included in eq.~(\ref{eMixedOrderModel}) for cosmetic reasons, the 
parameters $\lambda_m(w)$ are actually fixed to unity so that the 
model never looks further than $m$ words back.

We can view eq.~(\ref{eMixedOrderModel}) as a hidden variable
model. Imagine that we adopt the following strategy to predict the
word at time $t$.  Starting with the previous word, we toss a coin
(with bias $\lambda_1(w_{t-1})$) to see if this word has high
predictive value.  If the answer is yes, then we predict from the
skip-1 transition matrix, $M_1(w_{t-1},w_t)$.  Otherwise, we shift our
attention one word to the left and repeat the process.  If after $m-1$
tosses we have not settled on a prediction, then as a last resort, we
make a prediction using $M_m(w_{t-m},w_t)$.  The hidden variables in
this process are the outcomes of the coin tosses, which are unknown
for each word~$w_{t-k}$.

Viewing the model in this way, we can derive an EM algorithm to learn
the mixing coefficients $\lambda_k(w)$ and the transition
matrices\footnote{Note that the ML estimates of $M_k(w,w')$ do not
depend only on the raw counts of $k$-separated bigrams; they are also
coupled to the values of the mixing coefficients, $\lambda_k(w)$.  In
particular, the EM algorithm adapts the matrix elements to the
weighting of word combinations in eq.~(\ref{eMixedOrderModel}).  The
raw counts of $k$-separated bigrams, however, do give good initial
estimates.} $M_k(w,w')$.  The E-step of the algorithm is to compute,
for each word in the training set, the posterior probability that it
was generated by $M_k(w_{t-k},w_t)$.  Denoting these posterior
probabilities by $\phi_k(t)$, we have:
\begin{eqnarray}
\label{eMixedOrderEStep}
\lefteqn{\phi_k(t) =} \\
& & \frac{\lambda_k(w_{t-k}) M_k(w_{t-k},w_{t})
\prod_{j=1}^{k-1}[1\!-\!\lambda_j(w_{t-j})]}
{P(w_t|w_{t-1},w_{t-2},\ldots,w_{t-m})},\nonumber
\end{eqnarray}
where the denominator is given by eq.~(\ref{eMixedOrderModel}).  The
M-step of the algorithm is to update the parameters $\lambda_k(w)$ and
$M_k(w,w')$ to reflect the statistics in eq.~(\ref{eMixedOrderEStep}).
The updates for mixed-order Markov models are given by:
\begin{eqnarray}
\lambda_k(w) & \!\!\!\leftarrow \!\!\!&
\frac{\sum_t \delta(w,w_{t-k}) \phi_k(t)}
{\sum_t \sum_{j=k}^m \delta(w,w_{t-k}) \phi_j(t)},\\
M_k(w,w') & \!\!\!\leftarrow \!\!\!&
\frac{\sum_t \delta(w,w_{t-k})\delta(w',w_t) \phi_k(t)}
{\sum_t \delta(w,w_{t-k}) \phi_k(t)},
\end{eqnarray}
where the sums are over all the sentences in the training set, and
$\delta(w,w')=1$ iff $w=w'$.

\begin{figure}
\centerline{\hbox{\psfig{figure=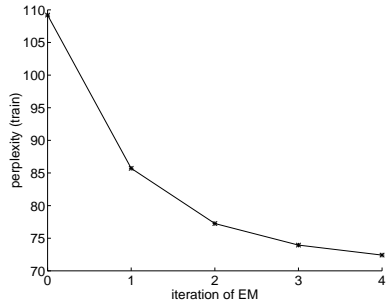,width=2in}}}
\caption{Plot of (training set) perplexity versus number of iterations
of the EM algorithm.  The results are for the $m=4$ mixed-order Markov
model.}
\label{fMixedOrderEM}
\end{figure}

We trained mixed-order Markov models with $2\le m \le 4$.
Figure~\ref{fMixedOrderEM} shows a typical plot of the training set
perplexity as a function of the number of iterations of the EM
algorithm.  Table~\ref{tMixedOrder} shows the final perplexities on
the training set (after four iterations of EM).  Mixed-order models
cannot be used directly on the test set because they predict zero
probability for unseen word combinations.  Unlike standard $n$-gram
models, however, the number of unseen word combinations actually {\it
decreases} with the order of the model.  The reason for this is that
mixed-order models assign finite probability to all $n$-grams $w_1 w_2
\ldots w_{n}$ for which {\it any} of the $k$-separated bigrams $w_k
w_{n}$ are observed in the training set.  To illustrate this point,
Table~\ref{tMixedOrder} shows the fraction of words in the test set
that were assigned zero probability by the mixed-order model.  As
expected, this fraction decreases monotonically with the number of
bigrams that are mixed into each prediction.

\begin{table}
\begin{center}
\begin{tabular}{|c|c|c|}
\hline
$m$ & train & missing \\
\hline
1 & 123.2 & 0.045 \\
2 & 89.4 & 0.014 \\
3 & 77.9 & 0.0063 \\
4 & 72.4 & 0.0037 \\
\hline
\end{tabular}
\end{center}
\caption{Results for ML mixed-order models; $m$ denotes the number
of bigrams that were mixed into each prediction.  The first column
shows the perplexities on the training set.  The second shows the
fraction of words in the test set that were assigned zero
probability. The case $m=1$ corresponds to a ML bigram model.}
\label{tMixedOrder}
\end{table}

Clearly, the success of mixed-order models depends on the ability to 
gauge the predictive value of each word, relative to earlier words in 
the same sentence.  Let us see how this plays out for the second-order 
($m=2$) model in Table~\ref{tMixedOrder}.  In this model, a small 
value for $\lambda_1(w)$ indicates that the word $w$ typically carries 
less information that the word that precedes it.  On the other hand, a 
large value for $\lambda_1(w)$ indicates that the word $w$ is highly 
predictive.  The ability to learn these relationships is confirmed by 
the results in Table~\ref{tLambda}.  Of the three-hundred most common 
words, Table~\ref{tLambda} shows the fifty with the lowest and highest 
values of $\lambda_1(w)$.  Note how low values of $\lambda_1(w)$ are 
associated with prepositions, mid-sentence punctuation marks, and 
conjunctions, while high values are associated with ``contentful'' 
words and end-of-sentence markers.  (A particularly interesting 
dichotomy arises for the two forms ``a'' and ``an'' of the indefinite 
article; the latter, because it always precedes a word that begins 
with a vowel, is inherently more predictive.)  These results 
underscore the importance of allowing the coefficients 
$\lambda_{1}(w)$ to depend on the context~$w$, as opposed to being 
context-independent (Ney, Essen, and Kneser, 1994).

\begin{table}
\begin{center}
{\footnotesize
\setlength{\listwidth}{2.8in}
\begin{tabular}{|c|}
\hline
$0.1 < \lambda_1(w) < 0.7 $ \\ \hline
\wordlist{\myquote{-} and of \myquote{"} or \myquote{;} to
\myquote{,} \myquote{\&} by with S. from nine were for that eight
low seven the \myquote{(} \myquote{:} six are not  against was four
between a their two three its \myquote{unknown} B. on as is
\myquote{---} 
five \myquote{)} into C. M. her him over than A.}
\\ \hline\hline
$0.96 < \lambda_1(w) \le 1$ \\ \hline
\wordlist{officials prices which go way he last they
earlier an
Tuesday there foreign quarter she former federal don't 
days Friday next Wednesday \myquote{\%} Thursday I Monday
Mr. we half based part United it's years going nineteen thousand
months \myquote{$\cdot$} million very cents San ago U. percent billion
\myquote{?} according \myquote{.} } \\ \hline
\end{tabular}
} 
\end{center}
\caption{Words with low and high values of $\lambda_1(w)$
in an $m=2$ mixed order model.}
\label{tLambda}
\end{table}

\section{Smoothing}
\label{sSmooth}
Smoothing plays an essential role in language models where ML
predictions are unreliable for rare events.  In $n$-gram modeling, it
is common to adopt a recursive strategy, smoothing bigrams by
unigrams, trigrams by bigrams, and so on.  Here we adopt a similar
strategy, using the $(m-1)$th mixed-order model to smooth the $m$th
one.  At the ``root'' of our smoothing procedure, however, lies not a
unigram model, but an aggregate Markov model with $C>1$ classes.  As
shown in Section~\ref{sAggregate}, these models assign finite
probability to all word combinations, even those that are not observed
in the training set.  Hence, they can legitimately replace unigrams as
the base model in the smoothing procedure.

Let us first examine the impact of replacing unigram models by
aggregate models at the root of the smoothing procedure.  To this end,
a held-out interpolation algorithm (Jelinek and Mercer, 1980) was used
to smooth an ML bigram model with the aggregate Markov models from
Section~\ref{sAggregate}. The smoothing parameters, one for each row
of the bigram transition matrix, were estimated from a validation set
the same size as the test set.  Table~\ref{tAggregateSmooth} gives the
final perplexities on the validation set, the test set, and the unseen
bigrams in the test set.  Note that smoothing with the $C=32$
aggregate Markov model has nearly halved the perplexity of unseen
bigrams, as compared to smoothing with the unigram model.

\begin{table}
\begin{center}
\begin{tabular}{|c|c|c|c|}
\hline
$C$ & validation & test & unseen \\ \hline
1 & 163.615 & 167.112 & 293175 \\
2 & 162.982 & 166.193 & 259360 \\
4 & 161.513 & 164.363 & 200067 \\
8 & 161.327 & 164.104 & 190178 \\
16 & 160.034 &162.686 & 164673 \\
32 & 159.247 & 161.683 & 150958 \\
\hline
\end{tabular}
\end{center}
\caption{Perplexities of bigram models
smoothed by aggregate Markov models with
different numbers of classes ($C$).}
\label{tAggregateSmooth}
\end{table}

Let us now examine the recursive use of mixed-order models to obtain
smoothed probability estimates.  Again, a held-out interpolation
algorithm was used to smooth the mixed-order Markov models from
Section~\ref{sMix}.  The $m$th mixed-order model had $mV$ smoothing
parameters $\sigma_{k}(w)$, corresponding to the $V$ rows in each
skip-$k$ transition matrix.  The $m$th mixed-order model was smoothed
by discounting the weight of each skip-$k$ prediction, then filling in
the leftover probability mass by a lower-order model.  In particular,
the discounted weight of the skip-$k$ prediction was given by
\begin{equation}
[1-\sigma_k(w_{t-k})]\lambda_k(w_{t-k})
\prod_{j=1}^{k-1}[1-\lambda_j(w_{t-j})]\quad\mbox{,}
\end{equation}
leaving a total mass of
\begin{equation}
\sum_{k=1}^m \sigma_k(w_{t-k}) \lambda_k(w_{t-k})
\prod_{j=1}^{k-1}[1-\lambda_j(w_{t-j})]
\end{equation}
for the $(m-1)$th mixed-order model.  (Note that the $m=1$ mixed-order
model corresponds to a ML bigram model.)

Table~\ref{tSmooth} shows the perplexities of the smoothed mixed-order
models on the validation and test sets.  An aggregate Markov model
with $C=32$ classes was used as the base model in the smoothing
procedure.  The first row corresponds to a bigram model smoothed by a
aggregate Markov model; the second row corresponds to an $m=2$
mixed-order model, smoothed by a ML bigram model, smoothed by an
aggregate Markov model; the third row corresponds to an $m=3$
mixed-order model, smoothed by a $m=2$ mixed-order model, smoothed by
a ML bigram model, etc.  A significant decrease in perplexity occurs
in moving to the smoothed $m=2$ mixed-order model.  On the other hand,
the difference in perplexity for higher values of $m$ is not very
dramatic.

\begin{table}
\begin{center}
\begin{tabular}{|c|c|c|}
\hline
$m$ & validation & test \\
\hline
1 & 160.1 & 161.3 \\
2 & 135.3 & 136.9 \\
3 & 131.4 & 133.5 \\
4 & 131.2 & 133.7 \\
\hline
\end{tabular}
\end{center}
\caption{Perplexities of smoothed mixed-order models
on the validation and test sets.}
\label{tSmooth}
\end{table}

Our last experiment looked at the smoothing of a trigram model.  Our
baseline was a ML trigram model that backed off \footnote{We used a
backoff procedure (instead of interpolation) to avoid the
estimation of trigram smoothing parameters.} to bigrams (and
when necessary, unigrams) using the Katz backoff procedure (Katz,
1987).  In this procedure, the predictions of the ML trigram model are
discounted by an amount determined by the Good-Turing coefficients;
the leftover probability mass is then filled in by the backoff
model. We compared this to a trigram model that backed off to the
$m=2$ model in Table~\ref{tSmooth}.  This was handled by a slight
variant of the Katz procedure (Dagan, Pereira, and Lee, 1994) in which
the mixed-order model substituted for the backoff model.

One advantage of this smoothing procedure is that it is 
straightforward to assess the performance of different backoff models.  
Because the backoff models are only consulted for unseen word 
combinations, the perplexity on these word combinations serves as a 
reasonable figure-of-merit.

Table~\ref{tUnseen} shows those perplexities for the two smoothed
trigram models (baseline and backoff).  The mixed-order smoothing was
found to reduce the perplexity of unseen word combinations by 51\%.
Also shown in the table are the perplexities on the entire test set.
The overall perplexity decreased by 16\%---a significant amount
considering that only 24\% of the predictions involved unseen word
combinations and required backing off from the trigram model.

\begin{table}
\begin{center}
\begin{tabular}{|c|c|c|}
\hline
backoff & test & unseen \\
\hline
baseline & 95.2 & 2799 \\
mixed & 79.8 & 1363 \\
\hline
\end{tabular}
\end{center}
\caption{Perplexities of two smoothed trigram models on the test set
and the subset of unseen word combinations.  The baseline model backed
off to bigrams and unigrams; the other backed off to the $m=2$ model
in Table~\protect\ref{tSmooth}.}
\label{tUnseen}
\end{table}

The models in Table~\ref{tUnseen} were constructed from all $n$-grams
($1\le n\le 3$) observed in the training data.  Because many $n$-grams
occur very infrequently, a natural question is whether {\em truncated}
models, which omit low-frequency $n$-grams from the training set, can
perform as well as untruncated ones.  The advantage of truncated
models is that they do not need to store nearly as many non-zero
parameters as untruncated models.  The results in
Table~\ref{tTruncated} were obtained by dropping trigrams that occurred
less than $t$ times in the training corpus.  The $t=1$ row corresponds
to the models in Table~\ref{tUnseen}.  The most interesting
observation from the table is that omitting very low-frequency
trigrams does not decrease the quality of the mixed-order model, and
may in fact slightly improve it.  This contrasts with the standard
backoff model, in which truncation causes significant increases in
perplexity.
\begin{table}
\begin{center}
\begin{tabular}{|c|c|c|c|c|}
\hline
$t$ & baseline & mixed & trigrams$(\times 10^6$) & missing \\
\hline
1 & 95.2 & 79.8 & 25.4 & 0.24 \\
2 & 98.6 & 78.3 & 6.1 & 0.32 \\
3 & 101.7 & 79.6 & 3.3 & 0.36 \\
4 & 104.2 & 81.1 & 2.3 & 0.38 \\
5 & 106.2 & 82.4 & 1.7 & 0.41 \\
\hline
\end{tabular}
\end{center}
\caption{Effect of truncating trigrams that occur less than $t$ times.
The table shows the baseline and mixed-order perplexities on the test
set, the number of distinct trigrams with $t$ or more counts, and the
fraction of trigrams in the test set that required backing off.}
\label{tTruncated}
\end{table}

\section{Discussion}
\label{sDiscussion}
Our results demonstrate the utility of language models that are
intermediate in size and accuracy between different order $n$-gram
models.  The two models considered in this paper were hidden variable
Markov models trained by EM algorithms for maximum likelihood
estimation.  Combinations of intermediate-order models were also
investigated by Rosenfeld (1996).  His experiments used the
20,000-word vocabulary {\em Wall Street Journal} corpus, a predecessor
of the NAB corpus.  He trained a maximum-entropy model
consisting of unigrams, bigrams, trigrams, skip-2 bigrams and
trigrams; after selecting long-distance bigrams (word triggers) on 38
million words, the model was tested on a held-out 325 thousand word
sample.  Rosenfeld reported a test-set perplexity of $86$, a $19\%$
reduction from the 105 perplexity of a baseline trigram backoff
model. In our experiments, the perplexity gain of the mixed-order
model ranged from 16\% to 22\%, depending on the amount of truncation
in the trigram model.

While Rosenfeld's results and ours are not directly comparable, both
demonstrate the utility of mixed-order models.  It is worth
discussing, however, the different approaches to combining information
from non-adjacent words.  Unlike the maximum entropy approach, which
allows one to combine many non-independent features, ours calls for a
careful Markovian decomposition.  Rosenfeld argues at length against
na\"{\i}ve linear combinations in favor of maximum entropy methods.
His arguments do not apply to our work for several reasons.  First, we
use a large number of context-dependent mixing parameters to optimize
the overall likelihood of the combined model.  Thus, the weighting in
eq.~(\ref{eMixedOrderModel}) ensures that the skip-$k$ predictions are
only invoked when the context is appropriate.  Second, we adjust the
predictions of the skip-$k$ transition matrices (by EM) so that they
match the contexts in which they are invoked.  Hence, the count-based
models are interpolated in a way that is ``consistent'' with their
eventual use.

Training efficiency is another issue in evaluating language models.
The maximum entropy method requires very long training times: e.g.,
200 CPU-days in Rosenfeld's experiments.  Our methods require
significantly less; for example, we trained the smoothed $m=2$
mixed-order model, from start to finish, in less than 12 CPU-hours
(while using a larger training corpus).  Even accounting for
differences in processor speed, this amounts to a significant mismatch
in overall training time.

In conclusion, let us mention some open problems for further research.
Aggregate Markov models can be viewed as approximating the full bigram
transition matrix by a matrix of lower rank.  (From
eq.~(\ref{eAggregate}), it should be clear that the rank of the
class-based transition matrix is bounded by the number of classes,
$C$.)  As such, there are interesting parallels between
Expectation-Maximization (EM), which minimizes the approximation error
as measured by the KL divergence, and singular value decomposition
(SVD), which minimizes the approximation error as measured by the
$L_2$ norm (Press {\em et al.}, 1988; Sch\"{u}tze, 1992).  Whereas SVD finds
a global minimum in its error measure, however, EM only finds a local
one.  It would clearly be desirable to improve our understanding of
this fundamental problem.

In this paper we have focused on bigram models, but the ideas and 
algorithms generalize in a straightforward way to higher-order 
$n$-grams.  Aggregate models based on higher-order $n$-grams (Brown 
{\em et al.}, 1992) might be able to capture multi-word structures 
such as noun phrases.  Likewise, trigram-based mixed-order models 
would be useful complements to $4$-gram and $5$-gram models, which are 
not uncommon in large-vocabulary language modeling.

A final issue that needs to be addressed is scaling---that is, how the
performance of these models depends on the vocabulary size and amount
of training data.  Generally, one expects that the sparser the data,
the more helpful are models that can intervene between different order
$n$-grams.  Nevertheless, it would be interesting to see exactly how
this relationship plays out for aggregate and mixed-order Markov
models.

\section*{Acknowledgments}
We thank Michael Kearns and Yoram Singer for useful discussions, the
anonymous reviewers for questions and suggestions that helped improve
the paper, and Don Hindle for help with his language modeling tools,
which we used to build the baseline models considered in the paper.

\section*{References}

\vskip 0.1in\noindent
P. Brown, V. Della Pietra, P. deSouza, J. Lai, and R. Mercer.
1992.  Class-based $n$-gram models of natural language.
{\it Computational Linguistics} 18(4):467--479.

\vskip 0.1in\noindent
S. Chen and J. Goodman. 1996.  An empirical study of smoothing
techniques for language modeling.  In {\it Proceedings of the 34th
Meeting of the Association for Computational Linguistics}.

\vskip 0.1in\noindent
I. Dagan, F. Pereira, and L. Lee.  1994.  Similarity-based estimation
of word co-occurrence probabilities.  In {\it Proceedings of the 32nd
Annual Meeting of the Association for Computational Linguistics}.

\vskip 0.1in\noindent
A. Dempster, N. Laird, and D. Rubin. 1977. Maximum likelihood from
incomplete data via the EM algorithm.  {\it Journal of the Royal
Statistical Society} B39:1--38.

\vskip 0.1in\noindent
X. Huang, F. Alleva, H. Hon, M.-Y. Hwang, K.-F. Lee, and
R. Rosenfeld. 1993. The {\sc sphinx-ii} speech recognition system: an
overview. {\em Computer Speech and Language}, 2:137--148.

\vskip 0.1in\noindent
F. Jelinek and R. Mercer. 1980.  Interpolated estimation of Markov
source parameters from sparse data.  In {\it Proceedings of the
Workshop on Pattern Recognition in Practice}.

\vskip 0.1in\noindent
F. Jelinek, R. Mercer, and S. Roukos.  1992.  Principles of lexical
language modeling for speech recognition.  In S. Furui and M.
Sondhi, eds. {\it Advances in Speech Signal Processing}.  Mercer
Dekker, Inc.

\vskip 0.1in\noindent
S. Katz. 1987.  Estimation of probabilities from sparse data
for the language model component of a speech recognizer.
{\it IEEE Transactions on ASSP} 35(3):400--401.

\vskip 0.1in\noindent
H. Ney, U. Essen, and R. Kneser.  1994.  On structuring probabilistic
dependences in stochastic language modeling.  {\it Computer
Speech and Language} {\bf 8}:1--38.

\vskip 0.1in\noindent
F. Pereira, N. Tishby, and L. Lee.  1993.  Distributional
clustering of English words.  In {\it Proceedings of the 30th
Annual Meeting of the Association for Computational Linguistics.}

\vskip 0.1in\noindent
W. Press, B. Flannery, S. Teukolsky, and W. Vetterling.  1988.  {\it
Numerical Recipes in C}.  Cambridge University Press: Cambridge.

\vskip 0.1in\noindent
R. Rosenfeld. 1996. A Maximum Entropy Approach to Adaptive Statistical
Language Modeling. {\em Computer Speech and Language}, 10:187--228.

\vskip 0.1in\noindent
H. Sch\"{u}tze. 1992. Dimensions of Meaning. In {\em Proceedings of
Supercomputing}, 787--796. Minneapolis MN.

\vskip 0.1in\noindent
Y. Singer.  1996.  Adaptive Mixtures of Probabilistic Transducers. In D.
Touretzky, M. Mozer, and M. Hasselmo (eds). {\it Advances in Neural Information
Processing Systems} 8:381--387. MIT Press: Cambridge, MA.

\end{document}